# Dynamic Decision Support System Based on Bayesian Networks

## Application to fight against the Nosocomial Infections

Hela Ltifi - Ghada Trabelsi - Mounir Ben Ayed - Adel M. Alimi

REGIM: REsearch Group on Intelligent Machines,
University of Sfax, National School of Engineers (ENIS),
BP 1173, Sfax, 3038, Tunisia

*Abstract*—The improvement of medical care quality is a significant interest for the future years. The fight against nosocomial infections (NI) in the intensive care units (ICU) is a good example. We will focus on a set of observations which reflect the dynamic aspect of the decision, result of the application of a Medical Decision Support System (MDSS). This system has to make dynamic decision on temporal data. We use dynamic Bayesian network (DBN) to model this dynamic process. It is a temporal reasoning within a real-time environment; we are interested in the Dynamic Decision Support Systems in healthcare domain (MDDSS).

## I. INTRODUCTION

The questions that interest health scientists become increasingly complex. For many questions, we need much time of analysis to generate significant quantities of complex temporal data that describe the interrelated histories of people and groups of people [8]. In Intensive Care Units (ICUs), physicians focus on the continuously evolution of patients. The temporal dimension plays a critical role in understanding the patients' state.

The development of methods for the acquisition, modeling and reasoning is, therefore, useful to exploit the large amount of temporal data recorded daily in the ICU. In this context, a Medical Decision Support System (MDSS) can be developed to help physicians to better understand the patient's temporal evolution in the ICU and thus to take decisions.

In many cases, the MDSS deals with the decision problem according to its knowledge; some of this knowledge can be extracted using a decision support tool which is the Knowledge Discovery from Databases (KDD) [10] [14]. The goal of the KDD is to extract knowledge and to interpret, evaluate and put it as a valid element of decision support.

The MDSS is well applied particularly to the prediction and shows significantly positive results in practice [7]. The control of the Nosocomial1 infections (NI) is regarded as a promising research field in the ICU [16]. These infections are contracted during the hospitalization.

From this point of view, a KDD-based MDSS aims at helping the physicians, users of the system, to especially understand and prevent the NI. The MDSS for the fight against NI require temporal data analysis. The dynamic aspect of the decisions is related to the measurements recorded periodically such as the infectious examinations, the antibiotic prescribed before admission, etc.

The objective is to daily predict the probability of acquiring a NI in order to daily follow-up the patient state using a KDD technique. With this intention, the data base must be pre-treated and transformed for a temporal data mining. The data mining technique must take into account the dynamic aspect of the decision. For this reason, we choose the Dynamic Bayesian Network (DBN) [9] [35] which are models representing uncertain knowledge on complex phenomena within the framework of a dynamic process. It is a question of obtaining knowledge models which evolve with time.

This article is organized into five sections. In the second, we will present the theoretical background of our decisional context. In section 3, we will concentrate on our problematic which is the fight against the nosocomial infections. We will also discuss the dynamic aspect of the decision. In section 4, we will describe, the use of the Dynamic Bayesian Networks as a KDD technique for supporting the dynamic medical decision-making. Concerning section 5, we will expose some results obtained by the application of the DBN for fight against NI. Finally, a conclusion and several perspectives will be proposed.

## II. KDD-BASED MDSS: SOLUTION EXPLOITED FOR THE MEDICAL FIELD

The decision is often regarded as a situation of choice where several solutions are possible; among them one is "the best" [34]. To decide is to choose in a reasonable way an appropriate alternative; it is a question of making a decision during a complete process [40]. Decision support systems play an increasingly significant role in medical practice. While helping the physicians or other professionals of the medical field to make clinical decisions, the MDSS exert a growing influence on the process of care for improved health care [30]. Their impact should be intensified because of our increasing capacity to treat more data effectively [21].

The MDSS can help the physicians to organize, store, and extract medical knowledge in order to make decisions. This can decrease the medical costs by providing a more specific and more rapid diagnosis, by a more effective treatment of the drugs prescriptions, and by reducing the need for specialists' consultations [32]. Within this framework, we are interested in the MDSS allowing controlling the NI which constitute a significant challenge of modern medicine and which are considered as one of the most precise indicators of the care quality of the patients [11].

---

[1] The term "nosocomial" comes from the Greek word "nosokomeion" to indicate the hospital

In medical decision making, Knowledge Discovery from Databases (KDD) [10] [14] is critical. In fact, knowledge, which is hidden in patient records, is valuable to provide precise medical decisions such as the diagnosis and the treatments. Indeed, traditional tools of decision support (OLAP, Info-center, dashboard, ERP, etc.) leave the initiative to the user to choose the elements which he/she wants to observe or analyze. However, in the case of KDD, the system often takes the initiative to discover associations between data. It is then possible, in a certain manner, to predict the future, according to the past.

The KDD is an interactive and iterative process aiming at extracting new, useful, and valid knowledge from a mass of data. It proceeds in four phases [10] [20] (Fig. 1):

1)   Selection of the data having a relationship with the analysis requested in the base;

2)   Cleaning of the data in order to correct the inaccuracies or data errors and transformation of the data into a format which prepares them for mining;

3)   The data mining, application of one or more techniques (neural networks, bayesian networks, decision tree, etc.) to extract the interesting patterns. A variety of KDD techniques were developed in the last few years and applied to the medical field; and

4)   Evaluation of the result allowing estimating the quality of the discovered model. Once knowledge is extracted, it is a question of integrating it by setting up the model or its results in the decisional system.

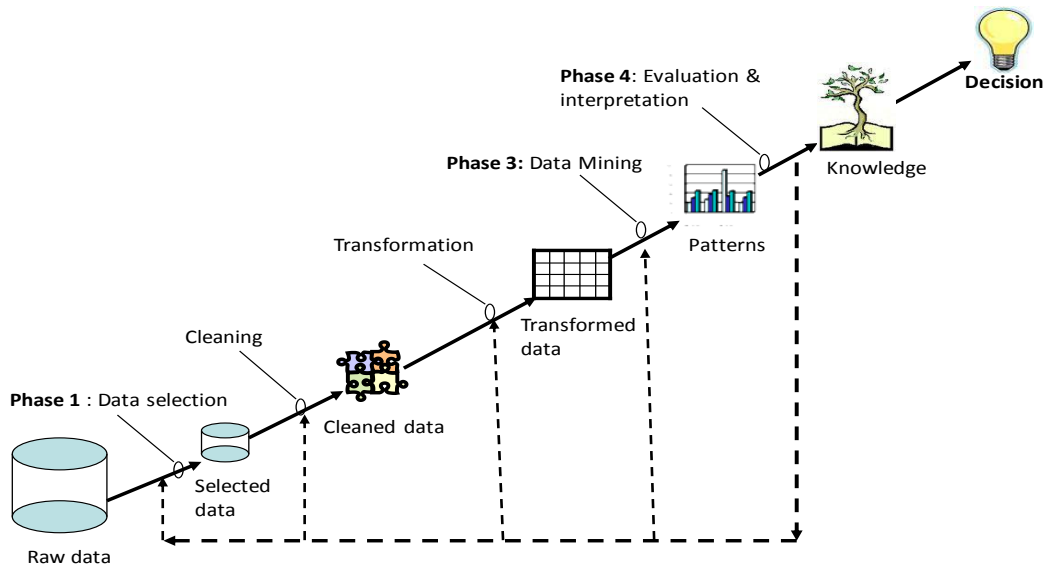

Figure 1.   KDD process

Various research tasks previously applied the assistance to the medical decision-making based on the KDD for the fight against NI [3] [4], there is no study (as far as we know) which addressed the dynamic aspect of the medical decision in this context.

### III.   DBN-BASED MDDSS: SOLUTION EXPLOITED FOR THE NI CONTROL

#### A.   Dynamic context

This article lies within the scope of a project aiming at fighting against NI[2] in the Intensive Care Unit in the Teaching hospital Habib Bourguiba in Sfax, Tunisia [1] [23] [24] [27] [26] [41]. Some work proposed NI control systems based on the KDD techniques [3] [4]. This Work shows their effectiveness and their capacity to produce useful rules. But, their direct use by doctors appears difficult to us. A study on a prevalence of NI occurrence in the Teaching hospital Habib Bourguiba in Sfax, Tunisia, showed that 17,9 % of the hospitalized patients were victims of a NI during 24 hours [16].

The decision problematic on the patient state must envisage and prevent the NI occurrence. The risks of this infection can weaken the patient or delay his cure. The risk of infection is mainly conditioned by the fragility of the patient and the ICU techniques used for its survival. Our objective is to predict the NI occurrence each day during the hospitalization period.

The dynamic aspect is observed on various levels of decision-making [38]. It is indispensable to take into account a set of critical factors of decision which are identified by the assistance in particular interviews with some of the ICU physicians. The identification of the factors supporting the appearance of the infections is a very significant stage which

---

[2] An infection is typically regarded as nosocomial if it appears 48 hours or more after hospital admission

influences the results of the decision-making. These factors are classified into two categories:

*1) Static data:* patient admission data (age, gender, weight, entry and exit dates, antecedents), the SAPS II3 score ([5] [31] proved that this score measured in the first 24 hours of the intensive care is an indicator of NI risk) and the Apache categorization 4 [18]. These data can help to determine the patients' fragility to the nosocomial infections.

*2) Temporal data:* control measurements to take each day (intubation, Central Venous Catheter (C.V.C.) [33], the urinary probe [13], Infectious examinations [16] [37] [42] and the catch of antibiotics.

At each day i ($1 \leq i \leq$ hospitalization duration), the decision on the patient state depends on the NI probability pi and thus on the values of the factors (static and temporal data) described above to the current day but also to the previous days, as well as to all the knowledge obtained by learning in time and recording former events. In fact, a basic decision is taken at the admission of the patient (t0). The future decision refers to a decision to be made after the consequences of a basic decision become (partially) known. A future decision is linked to the basic decision because the alternatives that will be available in the future depend on the choice made in the current basic decision. As time moves on, the future decision at current stage (t) becomes the basic decision at the next decision stage (t+1), when a new knowledge extracted by data mining (probability of acquiring a NI) and future decision should be addressed. This link repeats itself as long as the patient is hospitalized (cf. Fig. 2). The learn-then-decide-then-learn pattern describes how the decision-maker responds to new knowledge gained during the decision-making process. The elements described above, especially the existence of linked decisions, clearly show that decision-making in NI control is a dynamic process. In this scope, the decision-making process requires the consideration in time of linked or interdependent decisions, or decisions that influence each other. This dynamic decision-making pattern is a chain of decide, then learn; decide, then learn more; and so on. Such a system is so called Medical Dynamic Decision Support System (MDDSS).

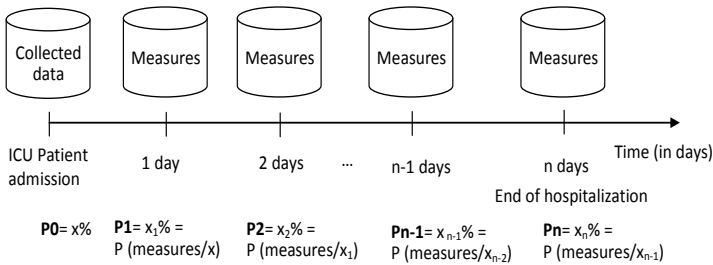

Figure 2. Temporal factors for the NI prevention

The MDDSS aims at the daily estimation of the NI occurrence probability, in percentage, during the ICU patient

---

hospitalization. This probability is calculated using a KDD technique. It is the content of the following section.

*B. Dynamic KDD technique*

Because of their capacity to represent uncertain knowledge, Bayesian networks (BN) play an increasingly important role in many medical applications. They have been introduced in the 1980s as a formalism of representation and reasoning with models of problems involving uncertainty and adopting probability theory as a basic framework. Research to explore the use of this formalism in the context of medical decision making started in the 1990s [36] [29].

The medical literature contains many examples of the BN use. We can quote a BN model developed to assist clinicians in the diagnosis and selection of antibiotic treatment for patients with pneumonia in the ICU [28]. Burnside and al. [6] proposed the use of BN to predict Breast Cancer Risk.

A BN is a Graphical model (marriage between probability theory and graph theory). It is a graph with probabilities for representing random variables and their dependencies. It efficiently encodes the joint probability distribution (JPD) of a set of variables. Its nodes represent random variables and its arcs represent dependencies between random variables with conditional probabilities. It is a directed acyclic graph (DAG) so that all edges are directed and there is no cycle when edge directions are followed [15] [19].

The joint probability distribution of random variables S = {$X_1, \ldots, X_N$} in a Bayesian network is calculated by the multiplication of the local conditional probabilities of all the nodes. Let a node $X_i$ in S denote the random variable $X_i$, and let $Pa(X_i)$ denote the parent nodes of $X_i$. Then, the joint probability distribution of S = {$X_1, \ldots, X_N$} is given by (1):

$$P(X_1, X_2, ..., X_N) = \prod_{i=1}^{N} p(X_i \mid Pa(X_i)) \qquad (1)$$

Unfortunately, a problem with the BN is that there is no mechanism for representing temporal relations between and within the random variables. For this reason, to represent variables that change over time, it is possible to use Dynamic Bayesian Networks (DBNs) [9] [35].

DBN encodes the joint probability distribution of a time-evolving set of variables X[t] = {$X_1[t], \ldots, X_N[t]$}. If we consider T time slices of variables, the dynamic Bayesian network can be considered as a "static" Bayesian network with T × N variables. Using the factorization property of Bayesian networks [9] [35], the joint probability density of $X_T$ = {$X[1], \ldots, X[T]$} can be written as (2):

$$P(X[1], ..., X[N]) = \prod_{t=1}^{T} \prod_{i=1}^{N} p(Xi[t] \mid Pa(X_i[t])) \quad \text{Where } Pa(X_i[t]) \text{ denotes the parents of } X_i[t] \quad (2)$$

DBNs are a generalization of Kalman Filter Models (KFM) and Hidden Markov Models (HMM). In the case of (HMM), the hidden state space can be represented in a factored form

---

instead of a single discrete variable. Usually dynamic Bayesian networks are defined using the assumption that X[t] is a first order Markov process [35] [39].

In the context of the NI prevention, the DBN technique uses fixed and temporal variables presented in the following section.

## IV. DBN APPLICATION FOR THE DATA MINING

### A. DBN variables

Our study concerns the application of the DBN technique on a temporal medical data base containing 280 patients' data.

The data acquisition and selection is the first KDD-based MDDSS phase. It concerns the implementation of the temporal data base that consists on a large collection of time series. It is a succession of couples $< (v_1,t_1),(v_2,t_2)\ldots,(v_i,t_i),\ldots>$ where $v_i$ is a value or a vector of values taken at a moment $t_i$. The values $v_i$ of a sequence are often real numbers [12]. In our context, the time series are a set of daily sequentially recorded values. The data pretreatment allows applying scripts to prepare useful variables for the knowledge extraction: (1) fixed data having only one value during the hospitalization period of a patient; and (2) temporal data having a value for each time serie (day) during the hospitalization period.

The estimation of the NI occurrence probability of the patient is represented by the following variables (table 1):

TABLE I. VARIABLES OF BAYESIAN NETWORKS

| Fixed variables | |
|---|---|
| **Code** | **Wording** |
| Sex | Patient gender |
| age1 | Patient age |
| Periode_entr | Indicates the entry season in ICU |
| Orig | Origin |
| Detorig | Origin details |
| priseAnti | Antibiotic catch |
| Knaus | Apache categorization of the previous patient state. |
| Cissue | Issue : the patient is dead or survived |
| Diag | Diagnosis |
| Ant | Antecedent |
| Result | Static NI prediction probability |
| **Temporal variables** | |
| **Code** | **Wording** |
| dsj | Difference between ICU admission and exit dates |
| acti | Act carried out at the day i |
| cissuei | Issue |
| examinfi | Infectious examinations at the day i |
| sensi | Sensibility to the germ (causing the Infectious examinations at the day i) to the prescribed antibiotic |
| resulti | dynamic NI prediction probability at day i |

The theory of the Bayesian Networks allows us to represent relationships between these observed variables in a probabilistic way which is well adapted to the uncertainty inherent to medical questions.

### B. Construction of knowledge model on fixed data (Static BN)

Causal links between the fixed variables are represented on figure 3. However observations made on a one static BN for a patient are not sufficient to estimate the NI occurrence probability.

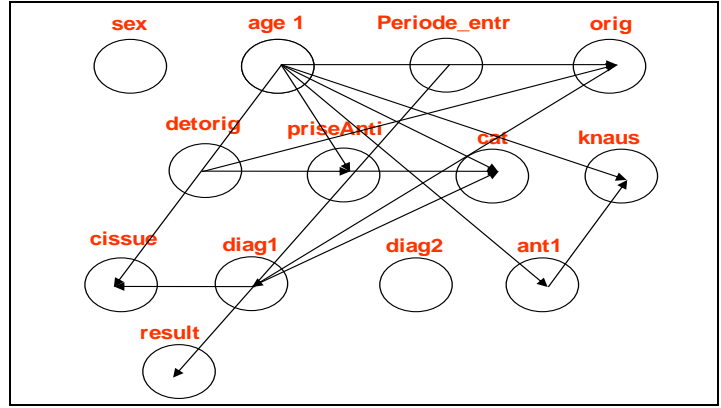

Figure 3. Causal links in static Bayesian network (static extracted knowledge model)

The extracted model could detect relations between logical variables like the relation between the age and the antecedent, between the age and cissue (the patient deceased or is survived). However obtained graph, contains "illogical" links between the nodes (for example, the age acts on the antibiotic catch). We also noted missing links which present interesting independence relations (For example, the relation between result and cissue).

The probabilities are calculated using $P(V_i|C)$ with:

- $V_i$ : the node (sex, age1, periode_entr… diag1) having discrete values, and

- C : the class to be predicted (cissue and result) having Boolean values (yes/no): the patient catches a NI or not

We obtain a static Bayesian Network: a causal graph with the probabilities associated to each node. The use of the probabilities and the causal graph provide knowledge models which are not very rich. So, experiments made with this BN showed that the prediction was instable and could produce false alerts. In order to represent the influence of past events over the present state of the patient, it is necessary to extend this model into a dynamic BN.

### C. Construction of knowledge model on temporal data (DBN)

The Figure 4 shows a dynamic extracted model based on temporal variables. The causal graph represents the interdependence between the temporal variables. We used for this dynamic structure the values of each time serie (act₁… act₁₀, exinf₁… exinf₃₀) [5] connected directly with the two predictive nodes which are the result and issue.

---

[5] In our context, we have 10 acts and 30 infectious examinations carried out daily to the patients.

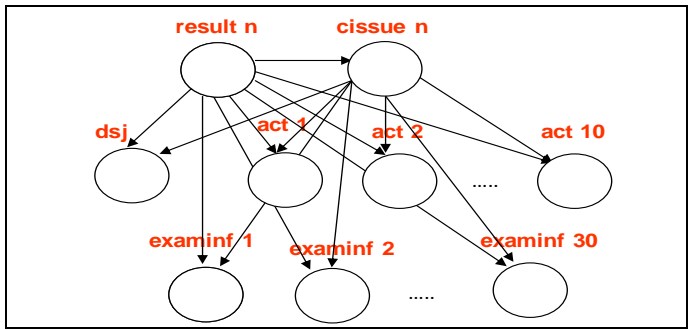

Figure 4.  Causal links in Dynamic Bayesian Network (extracted model for t=n)

The principle of our Dynamic Bayesian Network can be defined by:

- At t=0, we use extracted static knowledge model (figure 3)

- For $1 \leq t \leq T$ (patient hospitalization duration): unrolling the extracted temporal knowledge models (figure 4).

We obtain a final Dynamic Bayesian Network that has the following causal graph (figure 5).

The distribution result of the joined probabilities is given by (3):

$$P(\text{result}_{1:T}) = \prod_{t=1}^{T} \prod_{i=1}^{N} P(\text{result}_t^i | Pa(\text{result}_t^i)) \quad (3)$$

With:

- T is the interval of hospitalization time,

- N is the total number of the variables for each extracted model.

The DBN application gives good prediction results presented in the next section.

## V. PREDICTION RESULTS

This section presents the prediction results of an experimentation conducted over more than one year in the ICU of the teaching hospital Habib Bourguiba in Sfax, Tunisia.

After having generated many bases of examples, we applied our algorithm to real data coming from the ICU. We could extract knowledge models and transform them automatically to obtain probabilistic, quantitative and qualitative prediction results. These prediction results of our system are reliable to 74%, which is very encouraging.

Indeed, our study relates to the prediction of the patient state. This prediction is dynamic; it evolves throughout the patient hospitalization by new measurements.

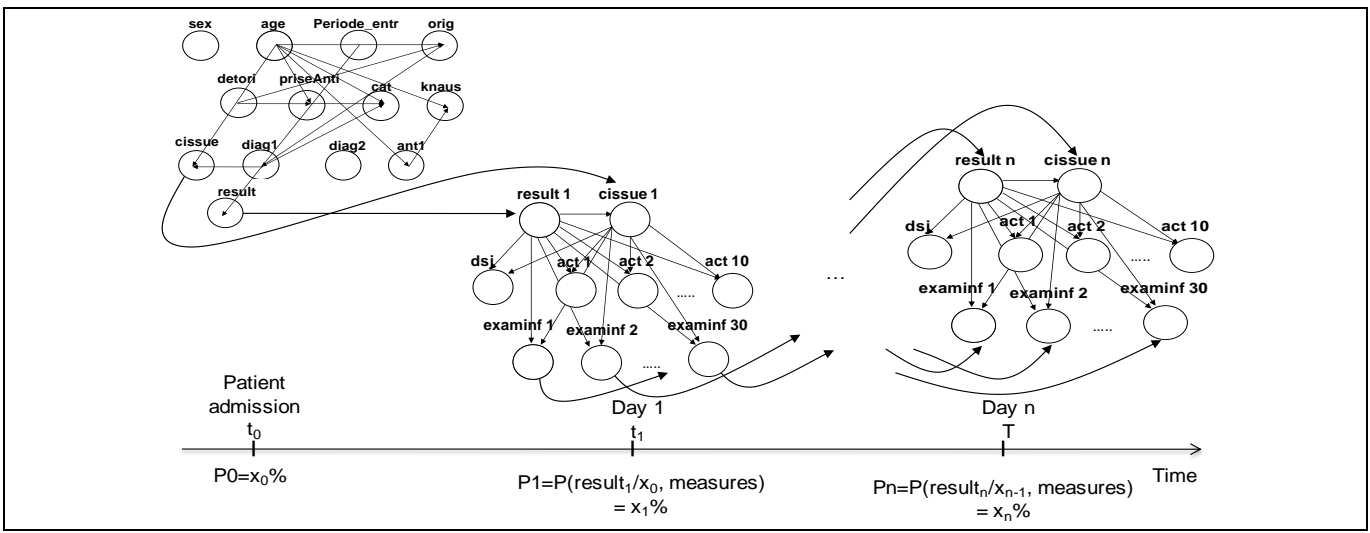

Figure 5.  The causal graph of the Dynamic Bayesian Network

With each dayi of the patient hospitalization, we could envisage his state at the future by a probability, which will be used, in the prediction of the dayi+1, with these measured observations.

We used a base of test which contains 58 cases (patients), for the performance evaluation of the system. We obtained the results given by the matrix of confusion[6] represented by the table 2.

TABLE II.  THE CONFUSION MATRIX OF THE RESULTS PROVIDED BY THE DYNAMIC BAYESIAN NETWORK

| | | Predicted | |
|---|---|---|---|
| | | *Negative* | *Positive* |
| **Actual** | *Negative* | 34 | 7 |
| | *Positive* | 8 | 9 |

We calculated the rates of evaluation starting from the prediction results obtained by our structure elaborated by the DBN. We found that the classification rate was correct to 0.74, the positive capacity of prediction = 0.56 and the negative

---

[6] Yes : to have a NI - No : not to have a NI - Total : the total of the predictions

capacity of prediction = 0.81. The generated observed vs. predicted results given by the table 2 are represented by the histogram (cf. Fig. 6).

An extension of the prediction phase could then be improved. With our current system, the prediction is made offline i.e. daily after the acquisition of all the data collected during the last 24 hours of hospitalization for a patient. This prediction can be improved so that it is carried out at each observation detected by our system and at every moment.

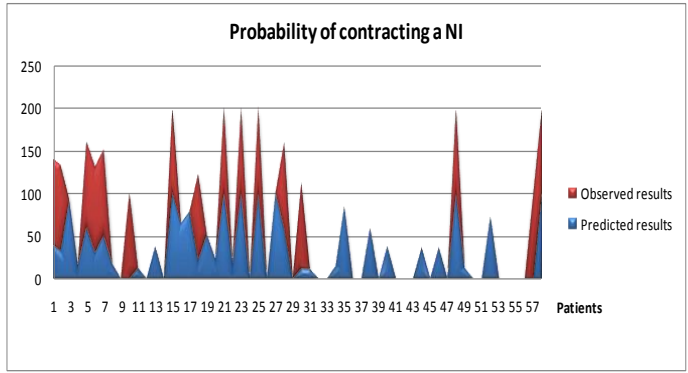

Figure 6.  Prediction results

## VI.  CONCLUSION

In this paper, we described an application of decision support system to the hospitalized patients in the ICU. This system aims at helping the physicians to estimate the NI appearance. The decision given by this system is dynamic because it is based on the patient state described in terms of a set of temporal factors of which the unit of time is the day. The dynamic decision system evolves and proceeds in several stages corresponding to the increasing levels of the patient situation comprehension (scale of time). On each level, a set of knowledge can be generated.

In this study we used the KDD as a decisional tool. A data pre-treatment is used in order to transform medical data into standardized data usable by the system. The KDD technique used is the Dynamic Bayesian Networks (DBN). It is used for the modeling of complex systems when the situations are dubious and/or the data are of complex structure. In our case, the complexity of the data is due to the fact that they are temporal and not regular.

We have implemented the dynamic BNs based on fixed (at t=0 that gives a static BN) and temporal data (daily taken measurements during the hospitalization stay). The application of the developed models for the NI prediction gives good results.

## VII.  FUTURE WORK

Under the angle of the Human-Computer Interaction (HCI) and basing on this experiment in the medical field, our research perspectives are related to the design and the evaluation of a MDDSS based on a KDD process. We are confronted to the need to develop a specific methodology for the design and the evaluation of DSS based on the KDD while taking starting point the criteria, methods and techniques resulting jointly from the HCI field [2] [22] and the visualization field [25]. This last technology makes it possible to present the data and knowledge in a visual form making it possible to the user to interpret the data, to draw the conclusions as well as to interact directly with these data. It is considered that the visualization techniques can improve the current KDD techniques by increasing the implication of the user and his confidence in connection with the observations discovered [17]. Such a methodology of evaluation must allow the study of cognitive and emotional experience of the DSS users for the fight against the nosocomial infections [43].

## ACKNOWLEDGMENT

The authors would like to acknowledge the financial support of this work by grants from General Direction of Scientific Research (DGRST), Tunisia, under the ARUB program. Thanks also to all the ICU staff of Habib Bourguiba Teaching Hospital for their interest to the project and all the time spent to help us design, use and evaluate our system.

## AUTHORS PROFILES

**Hela Ltifi profile**

Hela Ltifi has a Ph.D. in computer science. She is a member of the REGIM (ENIS, Tunisia). Her research activities concern DSS based on a KDD process and user-centred design; her application domain is hospital infections. She is a co-author of several papers in international journals and conferences. She is Assistant Professor in computer Sciences in the Faculty of Sciences of Gafsa, Tunisia. She is an IEEE member.

**Ghada Trabelsi  profile**

Ghada Trabelsi is a Ph.D. Student in Computer Sciences. She is a member of two laboratories: REGIM (ENIS, Tunisia) and LINA (Laboratoire d'Informatique de Nantes Atlantique). Her research topics concern Bayesian Networks, espesialy and there appliccation in the temporal and dynamic data. She is contractual Assistant in IPEIS (Institut Préparatoir aux Etudes d'Ingénieurs de Sfax). She is an IEEE member.

**Mounir Ben Ayed profile**

Mounir Ben Ayed is gradted  in Masrtery on Sciences and echnology. He obteined a PhD in Sciences - Biomedical Engineering. He is a member of the REGIM research unit, ENIS, University of Sfax, Tunisia. He is an assistant professor; he teaches DBMS, data warehouse and data mining. His research activities concern DSS based on a KDD process. Most of his research works are applied in health care domain. He is an author of many papers in international Journals and conferences. Mounir Ben Ayed has been a member of organization committees of several conferences, and particularly co-chair of the ACIDCA-ICMI'2005 organization comitee. He was the chair of the JRBA 2011(The 2nd Workshop on Bayesian Networks and their Applications) and he is the chair of the JFRB 2012 (The 6th French Speaker Workshop on Bayesian Networks). He is the former Director of Training at the Faculty of Sciences of Sfax. He is an IEEE member and chair  of IEEE-EMBS Tunisia Chapter

**Adel M. Alimi profile**

Adel M. Alimi is graduated in Electrical Engineering 1990, obtained a PhD and then an HDR both in Electrical & Computer Engineering in 1995 and 2000 respectively. He is now professor in Electrical & Computer Engineering at the University of Sfax. His research interest includes applications of intelligent methods (neural networks, fuzzylogic, evolutionary algorithms) to pattern recognition, robotic systems, vision systems, andindustrial processes. He focuses his research on intelligent pattern recognition, learning,analysis and intelligent control of large scale complex systems.

He is associate editor and member of the editorial board of many internationalscientific journals (e.g. "IEEE Trans. Fuzzy Systems", "Pattern RecognitionLetters", "NeuroComputing", "Neural Processing Letters", "International Journal of Imageand Graphics", "Neural Computing and Applications", "International Journal of Robotics andAutomation", "International Journal of Systems Science", etc.).

He was guest editor of several special issues of international journals (e.g. Fuzzy Sets& Systems, Soft Computing, Journal of Decision Systems, Integrated Computer AidedEngineering, Systems Analysis Modelling and Simulations).He is the Founder and Chair of many IEEE Chapter in Tunisia section, he is IEEE SfaxSubsection Chair (2011), IEEE ENIS Student Branch Counselor (2011), IEEE Systems,Man, and Cybernetics Society Tunisia Chapter Chair (2011), IEEE Computer Society TunisiaChapter Chair (2011), he is also Expert evaluator for the European Agency for Research.

He was the general chairman of the International Conference on Machine IntelligenceACIDCA-ICMI'2005 & 2000.He is an IEEE senior member.